\documentclass[letterpaper, 10 pt, conference]{ieeeconf}
\IEEEoverridecommandlockouts
\usepackage{cite}
\usepackage{amsmath,amssymb,amsfonts}
\usepackage{algorithm}
\usepackage{algpseudocode}
\usepackage{graphicx}
\usepackage{textcomp}
\usepackage{hyperref} 
\usepackage{xcolor}
\def\BibTeX{{\rm B\kern-.05em{\sc i\kern-.025em b}\kern-.08em
    T\kern-.1667em\lower.7ex\hbox{E}\kern-.125emX}}
\begin{document}
\newcommand{\bmat}[1]{\begin{bmatrix}#1\end{bmatrix}}

\title{Visual-Haptic Model Mediated Teleoperation for Remote Ultrasound}


\author{David Gregory Black$^{1,2}$, Maria Tirindelli$^{2}$, Septimiu Salcudean$^{1}$, Wolfgang Wein$^{2}$, and Marco Esposito$^{2}$
\thanks{$^{1}$Department of Electrical and Computer Engineering, University of British Columbia, Vancouver, Canada
        {\tt\small dgblack@ece.ubc.ca, tims@ece.ubc.ca}}%
\thanks{$^{2}$ImFusion GmbH, 80992 Munich, Germany
        {\tt\small tirindelli@imfusion.com, wein@imfusion.com, esposito@imfusion.com}}%
}

\maketitle

\begin{abstract}
Tele-ultrasound has the potential greatly to improve health equity for countless remote communities. However, practical scenarios involve potentially large time delays which cause current implementations of telerobotic ultrasound (US) to fail. Using a local model of the remote environment to provide haptics to the expert operator can decrease teleoperation instability, but the delayed visual feedback remains problematic. This paper introduces a robotic tele-US system in which the local model is not only haptic, but also visual, by re-slicing and rendering a pre-acquired US sweep in real time to provide the operator a preview of what the delayed image will resemble. A prototype system is presented and tested with 15 volunteer operators. It is found that visual-haptic model-mediated teleoperation (MMT) compensates completely for time delays up to 1000 ms round trip in terms of operator effort and completion time while conventional MMT does not. Visual-haptic MMT also significantly outperforms MMT for longer time delays in terms of motion accuracy and force control. This proof-of-concept study suggests that visual-haptic MMT may facilitate remote robotic tele-US. 
\end{abstract}


\section{Introduction}
With growing populations, global pandemics like COVID-19, and the rising economic and environmental cost of transportation, the ability to provide quality healthcare at a distance is of increasing importance. Furthermore, many countries struggle to provide adequate healthcare for their rural communities. For example, geographical isolation is cited as a central barrier to US (US) imaging for many remote communities in Canada \cite{adams2022}.

To solve this problem, a spectrum of solutions has been proposed and tested, from video conferencing-based tele-guidance to mixed reality ``human teleoperation" \cite{black2023hci}, to robotic teleoperation \cite{salcudean2000}. The latter has been heavily studied, with many recent surveys covering robotic US \cite{jiang2023,salcudean2022}, machine learning in robotic US \cite{bi2024}, and autonomous US scans \cite{li2021auto}.

Autonomous execution of a full US exam is still far from reality, so teleoperation is essential. In telerobotic US, an expert physician remotely controls a follower robot that has an US transducer on its end effector. The expert manipulates a master device to input their desired motion and receive force feedback through bilateral teleoperation. Sonographers usually look almost exclusively at the US image while scanning, relying on a haptic sense of the patient to move their transducer precisely. Moreover, it is often necessary to press relatively hard, commonly up to 25 N for abdominal scans \cite{black2024ft}, to modulate the force to avoid deforming organs, or to press in a certain way to look under the ribs or displace bowel gas that otherwise obscures the image. Without force feedback, such actions are not possible. A bilateral teleoperation system is transparent if the follower perfectly matches the master's trajectory and the master reflects exactly the force experienced by the follower.

\begin{figure*}
    \centering
    \includegraphics[width=0.95\linewidth]{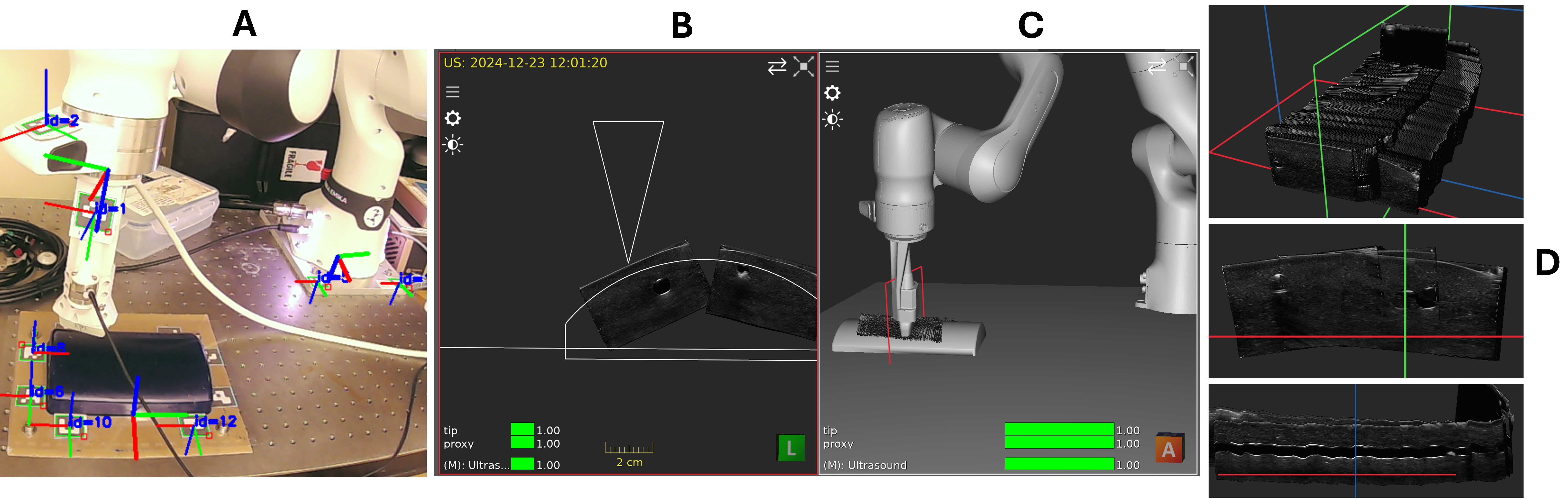}
    \caption{The local visual and haptic models shown to the expert during teleoperation. (A) shows the hand-eye and phantom calibration which allows the local visual and haptic models to be correctly co-located (C). The predicted US plane (red square in (C)) is used to re-slice the US sweep at any angle, generating the preview image in (B). Two orthogonal slices and a 3D view of the example sweep are shown in (D).}
    \label{fig:visualModel}
\end{figure*}
However, long-distance bilateral teleoperation over the Internet leads to potentially large and varying time delays between the master and follower, especially if the patient side is in a remote location with poor Internet connection. Even small time delays can quickly destabilize a nominally stable bilateral teleoperation system \cite{sheridan1993}. Many architectures have been proposed to overcome this challenge and guarantee stability despite delays, including wave variables \cite{niemeyer1991} and the time domain passivity approach (TDPA) \cite{hannaford2002}. While the former sacrifices tracking performance and transparency to achieve stability, TDPA can reach a high degree of transparency. However, the reflected force is still ultimately delayed, as are the video and US streams.

Some groups working on tele-US have implemented bilateral teleoperation with force feedback but negligible delay on a local network \cite{conti2014, mathiassen2016} while Arbeille et al. had a 1-2 second delay but no force feedback \cite{arbeille2018}. Fu et al. considered delays in their system and overcame them using model-mediated teleoperation (MMT) \cite{fu2022}.

In MMT, the remote environment is modeled and reproduced by the master as a virtual fixture, allowing the operator to interact with it directly \cite{mitra2008}. A virtual fixture is a surface or volume in space where a haptic device can apply an outward force when the user moves the end effector into the region, according to a virtual spring and damper or other impedance model \cite{ruspini1997}. The surface can be a mesh reconstructed from a depth or RGBD image of the environment, and the impedance can be estimated using force and pose sensing on the follower robot. By having the model locally, MMT gives the operator instant haptic feedback irrespective of time delay. As a result, it has been used successfully in space teleoperation \cite{yoon2004} and has shown better performance than TDPA in some low-velocity motions \cite{li2014}. The method depends on having a static or slowly varying environment with relatively constant mechanical impedance, although methods for near-real-time updates of the model and fast impedance estimation have been proposed \cite{xu2016}. The static assumption works relatively well for a patient undergoing an US exam.

However, instant haptic feedback does not account for the delay in the images. It has been shown that performance decreases when delay exceeds 150~ms for haptic tasks \cite{kaber2011}, but delays of only 69~ms in visual feedback are disruptive to a user manipulating a haptic device \cite{jay2005}. Furthermore, visual-haptic asynchrony occurs when visual and haptic signals are misaligned by more than 50-150 ms \cite{xu2016}. This can be disorienting and confusing. In tele-US specifically, the image is of utmost importance. Adams et al. stated that latency in the US video made it impossible to find the desired US plane in $24\%$ of their cases on a commercial system \cite{adams2022}, while others described teleoperated US with delayed images as ``very stressful" \cite{masuda2002}. Valenzuela-Urrutia et al. developed a virtual reality system that showed a virtual preview of the robot to give instant visual feedback in addition to point cloud-based haptics \cite{valenzuela2019}. However, this does not replace the US image, which is primarily what sonographers use to navigate.

To our knowledge, no research has addressed visual-haptic asynchrony in tele-US. This paper therefore proposes a new method of visual-haptic model-mediated tele-US in which not only the haptic model, but also an US model is recreated on the expert side. The expert thus receives immediate haptic and visual feedback and is able to carry out the US exam regardless of time delays. To this end, Section \ref{ssec:setup} introduces the concept and prototype system we have developed. Section \ref{ssec:exp} explains experiments that were carried out to quantify the utility of the method, with the results outlined in Section \ref{sec:res}.
\begin{figure*}[t]
    \centering
    \includegraphics[width=0.75\linewidth]{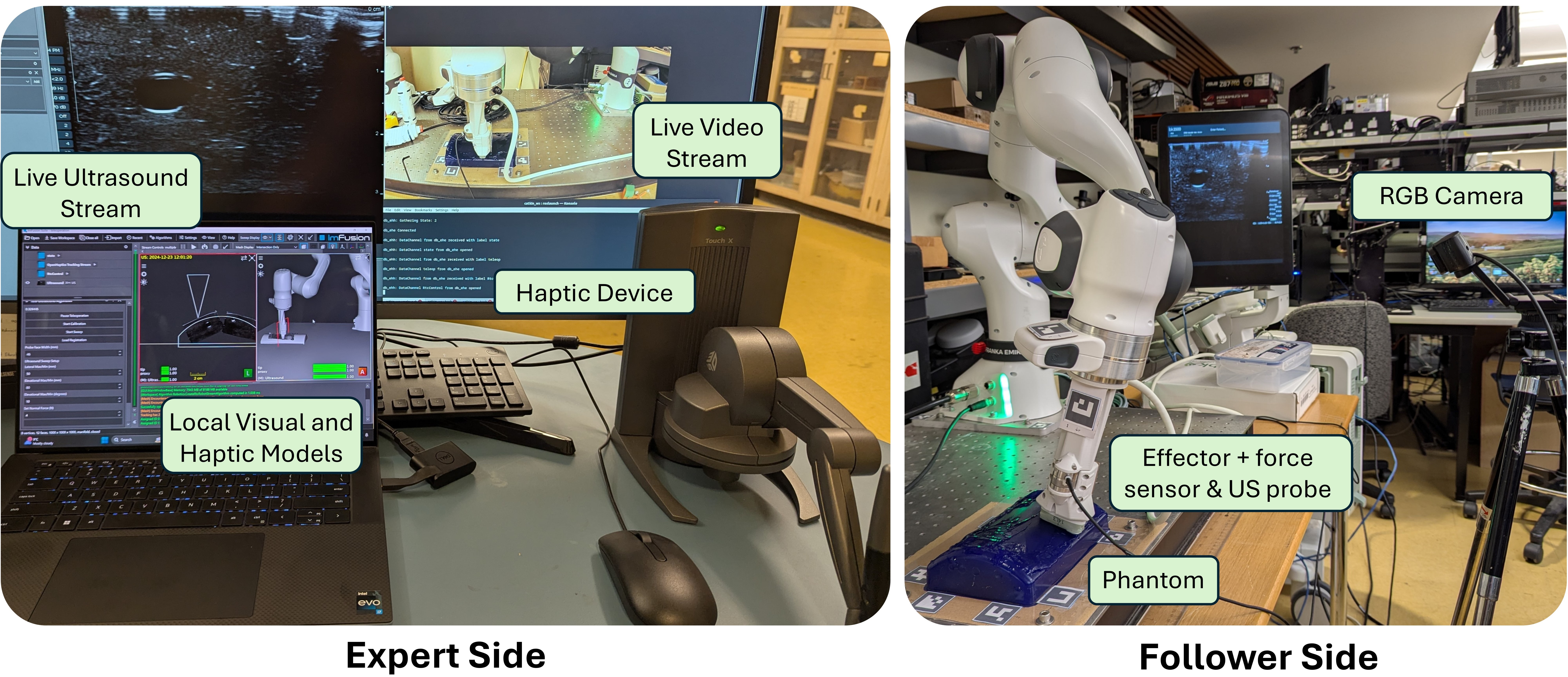}
    \caption{Overview of the visual-haptic MMT system. The ``live" streams here are those sent directly to the expert console, so they may be significantly delayed.}
    \label{fig:overview}
\end{figure*}

\section{Methods}\label{sec:meth}
\subsection{System Overview}\label{ssec:setup}
In visual-haptic MMT (VH-MMT), an RGB-D camera on the patient side captures the patient geometry, which is reconstructed on the expert side to create a virtual fixture for the master device. This is typical for MMT. In addition, however, the expert receives a local US model to enable delay-free visual feedback. Prior to the US exam, the follower robot completes an autonomous sweep of the patient's region of interest \cite{jiang2023, li2021auto}, which is transmitted to the expert and located spatially using the robot's pose from forward kinematics. The robot pose is hand-eye calibrated to the RGB-D camera using visual markers, and is calibrated to the US probe, so the pose of the sweep relative to the local haptic model is known. 

During the scan, when the expert moves their haptic device on the virtual fixture, the corresponding US plane is re-sliced from the pre-recorded sweep and displayed to the expert, thus providing instant visual feedback. A plane is computed according to the haptic device position and orientation, and the corresponding pixels from where the plane intersects the images in the sweep are interpolated to generate a new image closely approximating what the US probe would see in this pose on the real patient. This is achieved in real time through the fast GPU-based slicing implementation in the ImFusion Suite and is illustrated in Fig. \ref{fig:visualModel}. The visual feedback of the US slice is augmented by also showing the mesh of the patient and the URDF (unified robot description format) model of the follower robot, which preview the robot's motions. 

This method on its own may suffer from patient motion and dynamic factors such as bowel gas that are affected by the imaging itself. Therefore, the actual, delayed US image is also displayed to the expert. The expert can use the re-sliced image to achieve a correct rough positioning very efficiently, and subsequent slow, precise motions can use the real images to find the desired view. Since the fine motion is slower, it is less affected by the time delays. However, to ensure optimal performance, all incoming frames are integrated into the pre-recorded sweep to ensure it is up-to-date within the limits of the time delay. 

\subsection{Prototype System}
A prototype system to test the new MMT was developed using a collaborative, 7-degree-of-freedom (DoF) Panda robot (Franka Emika, Munich, Germany) with a custom US-probe-holding end effector and a BK Medical 14L3 linear US transducer, as shown in Fig. \ref{fig:overview}. A frame grabber was used to access the US image in real time. A RealSense RGB-D camera (Intel, Santa Clara, CA) was mounted facing the robot and patient from approximately $45^\circ$ above the horizontal. This was used for model reconstruction, and the RGB image was streamed to the expert. The expert side was implemented within the ImFusion Suite, including the standard US and mesh visualization and processing in addition to custom integrations of WebRTC for communication and OpenHaptics SDK to drive the haptic device. A Touch X haptic device (3dSystems, Rock Hill, SC) was used for the expert's input and force feedback.

\subsubsection{Robot Control}
The robot was controlled using ROS and LibFranka, as well as Ruckig \cite{berscheid2021} for trajectory smoothing and trac\_ik\cite{beeson2015} for inverse kinematics. LibFranka provided functions to obtain the configuration-dependent Coriolis, friction, and gravity forces in real time, as well as estimates of the mass matrix and Jacobian. We developed controllers taking these values as inputs, computing joint torques, and commanding them through the Franka Control Interface (FCI). To support high-rate control, the robot host computer was set up to run Linux 20.04 with a real-time kernel patch (PREEMPT\_RT). 

Two different custom robot controllers were tested, including jerk-limited joint impedance and Cartesian impedance. LibFranka's internal position and velocity controllers require continuous velocity and acceleration, as well as limited jerk. This makes it impossible to input the expert's desired pose directly as a set-point for the robot position/velocity controller. Instead, the trajectory must first be substantially interpolated and smoothed, which reduces the fidelity of the trajectory and increases lag. Furthermore, stiff position or rate control in the presence of a patient can be dangerous if the person is hit by the robot, or a large force is applied.

Instead, we first implemented a Cartesian impedance controller. This circumvents the jerk and acceleration limitations in the FCI by setting joint torques directly. It also provides a compliant interface for gentle and safe interaction with a patient, enabling the stiffness to be reduced in the direction normal to the patient surface. The impedance controller was implemented with gravity, friction, and Coriolis force compensation and proportional-derivative nullspace control using the dynamically consistent Jacobian pseudo-inverse of Khatib \cite{khatib1987}. However, it was found that the robot pose drifted relative to the haptic device, likely because the friction compensation was insufficient. Sudden orientation changes also sometimes led to unpredictable trajectories to reach the new pose.

For more direct control, therefore, we developed a joint impedance controller. This greatly improved the teleoperation but makes it impossible to adjust the controller stiffness according to the patient's surface normal. Instead, constant gain values were chosen to give reasonable performance everywhere in the workspace of interest. This achieves the desired compliant behavior while giving precise control.

The trac\_ik library is used to solve the inverse kinematics. The joint angles can then be interpolated through Ruckig with relatively high jerk and acceleration limits to obtain a smooth and safe but still responsive trajectory. Because the velocity is limited, the trajectory may fall behind the desired motion and start to drift. To avoid this, the current joint errors are added to the desired joint velocity with gain $K_e$ before computing the required joint torques. This effectively compensates for drift.

Since the manipulator is redundant, the inverse kinematics may have more than one solution. To avoid unexpected motions such as jumping between elbow-up and down configurations, the inverse kinematics solver finds solutions using multiple different algorithms, and the solution that minimizes the distance to the current joint configuration is selected \cite{beeson2015}. The methods include a stochastic Newton-Raphson iteration and sequential quadratic programming. With the trac\_ik solver, this can take approximately 2 ms, which is too slow to include directly in the robot control loop. Instead, the incoming poses from the communication are processed directly on one thread, and the resulting joint vectors are pushed to a fast, thread-safe, lock-less queue\footnote{\href{https://moodycamel.com/blog/2014/a-fast-general-purpose-lock-free-queue-for-c++}{MoodyCamel} lockless threadsafe queue}. The robot controller, running in a separate thread, monitors the queue for new commands and proceeds accordingly. Since the controller runs at a higher rate than the communication and inverse kinematics, the Ruckig interpolation is performed in the controller thread. This allows several steps of the interpolation to occur between every new command, thus obtaining a smooth trajectory without lagging behind the desired motion.

The final commanded joint configuration, $\pmb{q}_d$, is achieved using a PD controller with Coriolis and friction ($\pmb{\tau}_{cf}$), and gravity ($\pmb{\tau}_g$) compensation, which gives the desired spring-damper behavior:
\begin{equation}
    \pmb{\tau}_c = K_p(\pmb{q}_d-\pmb{q}_m)  + K_d(\dot{\pmb{q}}_d-\dot{\pmb{q}}_m) + \pmb{\tau}_{g} + \pmb{\tau}_{cf}
\end{equation}
The full process is illustrated in Algorithm \ref{alg:joint}.

\begin{algorithm}[h]
    \caption{Motion-limited joint impedance controller with drift compensation}\label{alg:joint}
    \begin{algorithmic}
        \State $\pmb{q}_m$ is the measured joint configuration (on both threads)
        \State\vspace{1ex}
        \State \textbf{Thread 1:}
        \State $T_g$ is the goal pose received from the master
        \State $\{\pmb{q}_{gi}\} \gets \text{InverseKinematics}(T_g)$ 
        \State $\pmb{q}_g\gets \underset{\pmb{q}}{\text{argmin}}|| \pmb{q}-\pmb{q}_m||_2^2,~\pmb{q}\in\{\pmb{q}_{gi}\}$
        \State Push $\pmb{q}_g$ to Queue
        \State\vspace{1ex}
        \State \textbf{Thread 2:}
        \State $\pmb{q}_g\gets \pmb{q}_{last}$
        \If{Queue has a new goal configuration, $\pmb{q}_{new}$}
        \State $\pmb{q}_g\gets\pmb{q}_{new}$
        \EndIf
        \State Set $\pmb{q}_g$ as the goal in Ruckig
        \State Compute the next update step, $\pmb{q}_n,~\dot{\pmb{q}}_n$ using Ruckig's smoothing and interpolation
        \State Set $\dot{\pmb{q}}_d\gets \dot{\pmb{q}}_n + K_e(\pmb{q}_n-\pmb{q}_m)$
        \State $\dot{\pmb{q}}_d \gets $ Saturate $\dot{\pmb{q}}_d$ within safe bounds
        \State $\pmb{q}_d\gets\pmb{q}_{last}+dt\cdot\dot{\pmb{q}}_d$ and $\pmb{q}_{last}\gets\pmb{q}_n$
        \State $\pmb{\tau} \gets K_p(\pmb{q}_d-\pmb{q}_m)  + K_d(\dot{\pmb{q}}_d-\dot{\pmb{q}}_m) + \pmb{\tau}_{g} + \pmb{\tau}_{cf}$
        \State\indent 
    \end{algorithmic}
\end{algorithm}

\subsubsection{Communication}
The communication between expert and follower occurs over Web Real Time Communication (WebRTC)\footnote{\href{https://github.com/paullouisageneau/libdatachannel}{LibDataChannel} WebRTC and WebSockets}, a peer-to-peer framework that enables fast and secure communication over the Internet \cite{webrtc,sredojev2015}. The pose and force data, as well as various control commands are sent over WebRTC data channels, which use stream control transport protocol (SCTP) and are configured to send as fast as possible without retransmission to avoid delays. The video streams are sent over media tracks, which use secure real-time transport protocol (SRTP) and H.264 encoding.

The desired poses are received with a timestamp from the expert over WebRTC. The timestamp is first checked to ensure the message is current, after which the pose value is checked to ensure it was not corrupted. Finally, the pose is passed on for further processing and enqueued for the control loop to access, as described in the previous subsection. The measured robot state is pushed to a separate queue from which it is in turn dequeued by a loop on a third thread, which encodes and sends the state to the expert.

\subsubsection{Calibration}
We define the following homogeneous transformation matrices:
\begin{itemize}
\item ${}^sT_r$: RGB-D to follower base (hand-eye calibration)
\item ${}^fT_u$: US probe to follower flange (US calibration)
\item ${}^sT_f$: Follower flange to base
\item ${}^mT_h$: Expert handle to master base (pose of expert hand)
\end{itemize}
During teleoperation, the expert sees the scene through the RGB-D camera, so the probe pose they perceive is ${}^rT_u={}^sT_r^{-1}{}^sT_f{}^fT_u$. Thus, the goal of the controller is to achieve ${}^mT_h={}^rT_u$, which aligns the directions in the haptic device and video stream. By substitution of these two equations, the expert's input is transformed to a follower controller input as ${}^sT_f{}={}^sT_r{}^mT_h{}^fT_u^{-1}$. Therefore, we require the matrices of both the depth camera (hand-eye calibration) and the US probe (robot-US calibration) relative to the follower robot. 

A final transform is also required because the follower and master are not initially aligned. When starting the teleoperation, the expert is asked to hold the haptic device in a central position and align the orientation manually with the current follower orientation, since the wrist of the Touch X haptic device is not actuated. During the alignment phase, the angular error is printed to the screen so the expert can see when they are approaching the correct pose. Upon sufficient rotational alignment, the current pose offset, $T_{off}$ from the robot flange to the expert's input is computed and applied to all subsequent commands to avoid a jump in the robot control. Later in the teleoperation, if the expert has reached the end of the haptic device's limited workspace, they can hold the stylus button to freeze the teleoperation and move back to a central location, thus re-indexing the control. In this situation, once the expert releases the indexing button, the position offset is updated so the follower robot remains stationary throughout the process.

In both cases, suppose the current offset is $T_{off}$, the pre-indexing pose is ${}^sT_{f0}$, and the post-indexing commanded pose is ${}^sT_{f1}$. To avoid a jump, we require ${}^sT_{f1} = {}^sT_{f0}$. Therefore, the new offset is set to $T_{off} \gets {}^sT_{f0}\left(T_{off}^{-1}{}^sT_{f1}\right)^{-1}$.
Thus, at last, the commanded follower pose is
\begin{equation}
{}^sT_f{}=T_{off}{}^sT_r{}^mT_h{}^fT_u^{-1}
\end{equation}

There are numerous robot-US calibration procedures, most of which are time consuming and involve manual intervention \cite{li2021calib}. We instead used the CAD model of the end-effector and US probe and checked the calibration with a simple pivot test. The two values corresponded to within 1 mm. To find the hand-eye calibration, ArUco markers were placed on the robot and end effector in known poses, ${}^fT_{ai}$, as shown in Fig.~\ref{fig:visualModel}. The markers are tracked by the camera using OpenCV (${}^rT_{ai}$), and thus a calibration transform is computed by solving ${}^sT_f{}^fT_{ai}={}^sT_r{}^rT_{ai}$ for ${}^sT_r$. For simplicity, the separable solution of Zhuang et al. was used \cite{zhuang1994}. To handle the switching between alignment, calibration, indexing, and normal teleoperation, the follower robot controller runs in a finite state machine with state transitions triggered by button presses and alignment of the haptic device.

\begin{figure}[t]
    \centering
    \includegraphics[width=\linewidth]{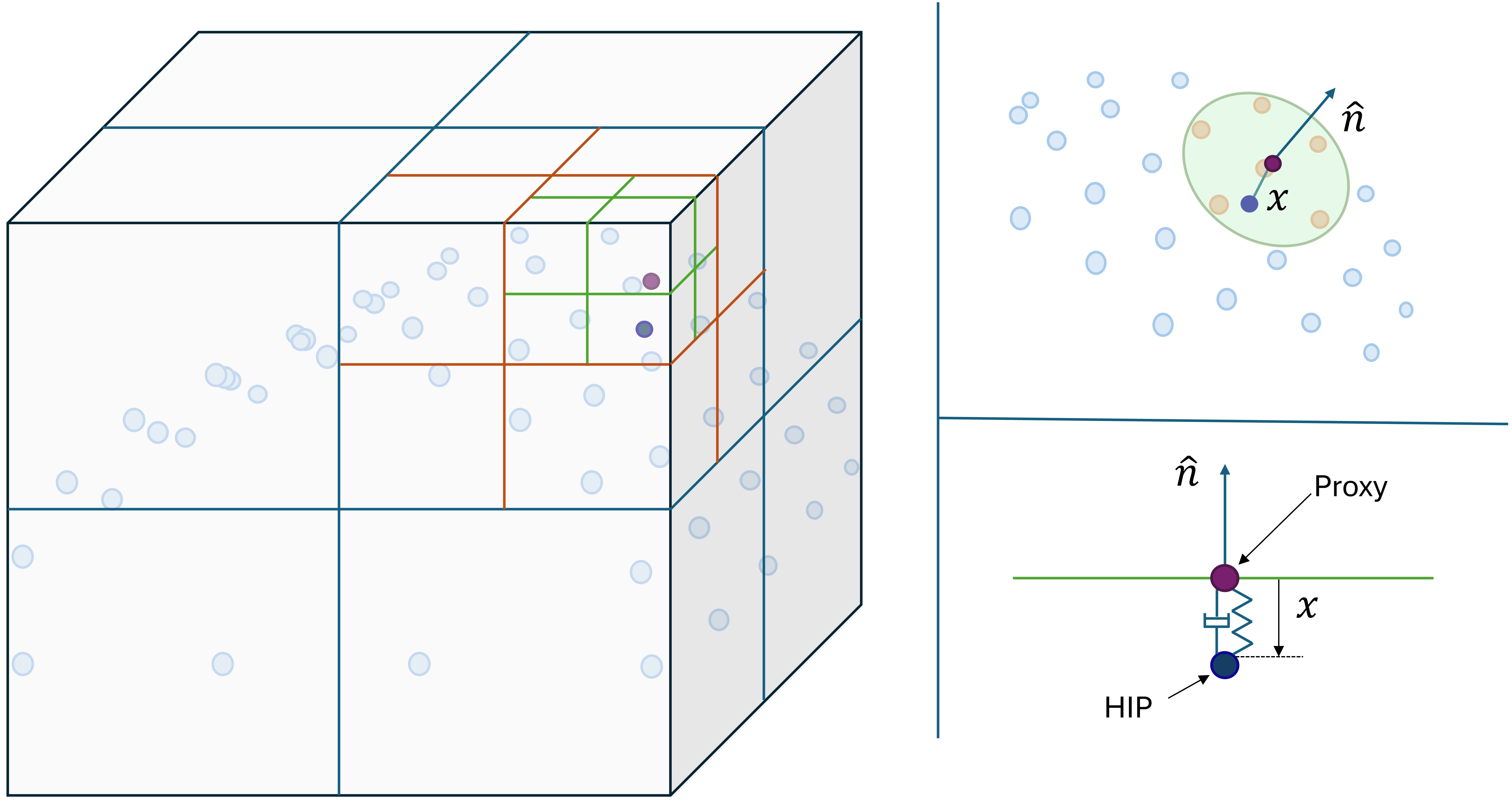}
    \caption{Point cloud-based haptics for model-mediated teleoperation. The cube represents an octree, which is used to search the point cloud. A surface is fitted to a neighborhood of points (green circle), and a virtual spring-damper is used to compute the force.\vspace{-2ex}}
    \label{fig:haptics}
\end{figure}
\subsubsection{Haptics}
Haptic feedback to the expert is achieved through a virtual fixture based on the measured point cloud of the patient from the RGB-D camera. The ImFusion Suite was used to compound many depth frames into a single detailed, relatively smooth point cloud representation of the environment. This point cloud was represented as an octree, with each node constituting an octant of the Euclidean space, and branching accordingly, as shown in Fig. \ref{fig:haptics}. This allowed for very efficient searching of the closest $n$ points to the haptic interface point (HIP). With this representation, we used the proxy point method for point clouds, in which a plane is fitted to a neighborhood of $n$ points around the HIP to define a surface and its normal \cite{ryden2011}. A virtual spring-damper between the HIP and a proxy point left on the surface is used to compute the force applied by the haptic device to the expert's hand. Since the RGB-D camera is co-registered to the follower robot, when the expert touches the virtual object, the robot touches the real one. The compliance of the virtual model and the robot controller ensure that this is true despite imperfections in the registration and reconstruction.

\subsection{Experiments}\label{ssec:exp}
To test the visual-haptic MMT compared to conventional MMT, we built the prototype system described above. For repeatability, we scanned a Blue Phantom branched 2-vessel training block (CAE Healthcare, Inc.). This was secured to the same optical prototyping board as the Franka robot. The US sweep and 3D model were measured once prior to testing and saved in the ImFusion Suite with the correct transform relative to the follower robot for time-saving and repeatability.

For the tests, volunteers were recruited to perform an US scanning task five times: VH-MMT and short (500 ms) communication delay, VH-MMT and long (1000 ms) delay, once for each delay with standard MMT, and once with no delay and standard MMT as a control. The order of the VH-MMT versus standard MMT was randomized to avoid learning effects, and the volunteers performed a practice scan prior to testing to become familiar with the interface. As shown in Fig. \ref{fig:overview}, the visual-haptic aspect was shown on a laptop screen, while the normal US and video streams were shown on an additional monitor. During conventional MMT tasks, the laptop screen was simply obscured from view.

To simulate the communication delays, all incoming/outgoing messages on the expert side were stored in queues along with their arrival/sending timestamp. A separate thread monitored the queues and sent the next item only when the desired time delay had elapsed. The stated time delays are single-direction delays and the round-trip times (RTTs) are approximately double the given values, i.e. 1 and 2 seconds. The communication system without time delays was measured to have an RTT of approximately 1 ms.

\begin{figure}[t]
    \centering
    \includegraphics[width=\linewidth]{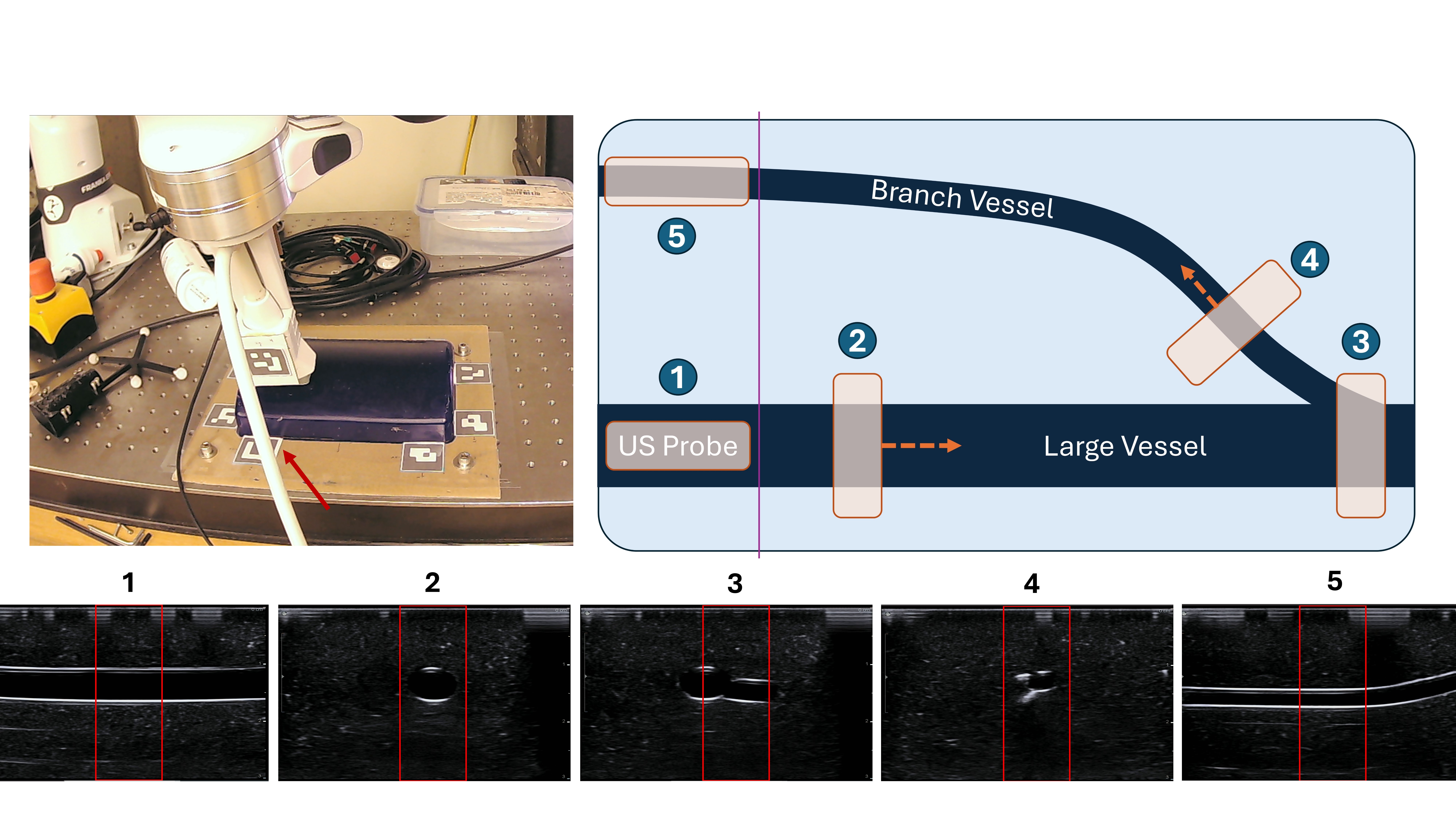}
    \caption{Plan of US scanning task on branching vessel phantom. Steps 1-5 are outlined in the text, and each view is shown in the US below. The red lines are guides for the user to keep the vessels centered. The ArUco marker edge indicated by the red arrow shows where the user must switch between the longitudinal and transverse views.\vspace{-2ex}}
    \label{fig:testPlan}
\end{figure}

The US scanning task consisted of the following steps, shown also in Fig. \ref{fig:testPlan}. (1) Find a clean longitudinal view of the left side of the large vessel. (2) With a transverse view, sweep along the length of the large vessel, keeping it centered. (3) Center the bifurcation of the two vessels. (4) Sweep along the thin vessel, keeping it centered. (5) Find a longitudinal view of the left side of the thin branch vessel.

A line on the phantom's board indicates the minimum extent of the sweeps, and the point to the left of which the longitudinal views must be captured. During the test, the screen showing the US image and the RGB stream was recorded, and red lines were rendered on the US image to indicate where the user should align the transverse views.

After the tests, the recorded images were evaluated for how well the subject kept the vessels centered and how consistent the applied force was. To do so, the vessel was segmented as an ellipse, and its centroid and semi-axis lengths, $a,~b$ were saved for each frame of the sweep. The eccentricity of the vessel, given by $e=\sqrt{1-\frac{a^2}{b^2}}$, was computed for each frame and is proportional to the applied force. Finally, the completion time was recorded for each sub-task and the subjects filled in modified NASA task load index (TLX) \cite{hart1988} forms after every test to score mental demand, physical demand, effort, and perceived performance on a scale of 1 to 10. 

\section{Results}\label{sec:res}
In total, 15 subjects completed the tests, performing 5 US scans each. From user feedback, it was much easier to find the bifurcation and longitudinal views of the vessels with VH-MMT than without the preview image. As a result, the completion times were significantly shorter with VH-MMT than with conventional MMT, as shown in Table~\ref{tab:telerob:completion}. While sweeping, the users carefully moved straight, no matter the delay, so the timing differences were small. Conversely, when finding an anatomy initially, the preview image was very useful, leading to significant differences. Additionally, the mental demand and effort were significantly lower and the perceived performance was higher, as shown in Table~\ref{tab:telerob:tlx}. Interestingly, none of the effort or completion time values differed significantly between 0 delay and 500 ms delay with VH-MMT, showing that VH-MMT compensated completely for the smaller delay in terms of user effort. Table~\ref{tab:telerob:mmtLateral} shows that the users were able to follow the vessels more accurately using VH-MMT at larger time delays, and Table~\ref{tab:telerob:mmtCompression} shows significantly more consistent pressure using VH-MMT. For most of these factors, the difference in performance between conventional and VH-MMT was larger with a more time delay.

\begin{table}[h]
\centering
\caption{Completion times (seconds) using conventional and VH MMT at different levels of communication delay. VH-MMT decreases completion time significantly.}
\begin{tabular}{|c|c|c|c|c|}
\hline
Delay & Method & Vessel Finding & Sweeping & Total\\\hline
0 ms &  MMT & $89\pm133$ & $99\pm35$ & $189\pm148$ \\\hline
500 ms & MMT & $128\pm80$ & $92\pm29$ & $220\pm90$ \\\hline
500 ms & VH-MMT & $69\pm44$ & $94\pm63$ & $163\pm97$\\\hline
1000 ms & MMT & $102\pm69$ & $89\pm57$ & $192\pm124$ \\\hline
1000 ms & VH-MMT & $83\pm79$ & $76\pm21$ & $159\pm96$ \\\hline
\multicolumn{2}{|c|}{p-Values - 500 ms} & 0.040 & 0.47 & 0.027 \\\hline
\multicolumn{2}{|c|}{p-Values - 1000 ms} & 0.013 & 0.07 &  0.022 \\\hline
\end{tabular}
\label{tab:telerob:completion}
\end{table}
\begin{table}[h!]
\centering
\caption{Results from NASA TLX questionnaire, showing that delays make the teleoperation more difficult, but VH-MMT significantly decreases mental demand and effort and increases perceived performance. Scores are 1 = least and 10 = most. Physical demand was approximately constant ($3.0\pm 1.9$).}
\begin{tabular}{|c|c|c|c|c|}\hline
Delay & Method & Mental Dem. & Effort & Performance\\\hline
0 ms & MMT & $3.5\pm1.2$ & $4.3\pm1.9$ & $7.8\pm1.4$\\\hline
500 ms & MMT & $5.6\pm1.5$ & $6.1\pm1.6$  & $5.8\pm0.9$\\\hline
500 ms & VH-MMT & $3.8\pm0.9$ & $4.3\pm1.6$ & $7.2\pm0.8$\\\hline
1000 ms & MMT & $7.8\pm0.8$ & $7.4\pm1.2$ & $4.4\pm0.9$\\\hline
1000 ms & VH-MMT & $5.6\pm0.8$ & $5.3\pm1.9$ &$6.1\pm0.6$ \\\hline
\multicolumn{2}{|c|}{p-Values - 500 ms} & 0.012 & 0.047 & 0.012\\\hline
\multicolumn{2}{|c|}{p-Values - 1000 ms} & 0.018 & 0.019 &  0.003\\\hline
\end{tabular}
\label{tab:telerob:tlx}
\end{table}

\begin{table}[h!]
\centering
\caption{Lateral root mean square error (RMSE; pixels) of vessel following. The subjects aimed to keep the vessels centered. VH-MMT made it easier to achieve this goal.}
\begin{tabular}{|c|c|c|}\hline
Type & Offset (500ms Delay) & Offset (1000ms Delay)\\\hline
MMT &  $231.4$ & $234.2$\\\hline
VH-MMT & $161.0$ & $180.7$\\\hline
p-Value & $<0.001$ & $0.031$\\\hline
\end{tabular}
\label{tab:telerob:mmtLateral}
\end{table}

\begin{table}[h!]
\centering
\caption{Vessel eccentricity during the sweeps. Larger eccentricity shows excessive force, and larger standard deviation shows less consistent application of pressure. VH-MMT led to better consistency and slightly better compression.}
\begin{tabular}{|c|c|c|c|}\hline
Type & MMT & VH-MMT & P-Value \\\hline
Large Vessel (500ms) &  $0.20\pm0.13$ & $0.12\pm0.09$ & $<0.001$\\\hline
Branch Vessel (500ms) &  $0.46\pm0.25$ &  $0.33\pm0.19$ &$<0.001$ \\\hline
Large Vessel (1000ms) & $0.24\pm0.17$ & $0.20\pm0.13$ & $<0.001$\\\hline
Branch Vessel (1000ms) & $0.43\pm0.28$ & $0.41\pm0.23$ & $0.03$\\\hline
\end{tabular}
\label{tab:telerob:mmtCompression}
\end{table}


\section{Discussion and Conclusion}
In every category from mental demand and effort, perceived performance, and completion time to motion accuracy and force consistency, visual-haptic MMT outperformed conventional MMT significantly. In terms of effort and completion time, VH-MMT completely compensated for a 500 ms teleoperation delay, leading to performance not significantly different from the zero time delay case. We found that the VH-MMT is particularly useful to find the correct rough position when the operator is not already in the close vicinity of a desired view. In the above tests, however, the operator was usually already close to the desired position at the end of the sweeps. Thus, the positive effect of VH-MMT here may be underestimated.

On the other hand, the primary limitation of visual-haptic MMT is that the model is relatively static while the patient may move, breathe deeply, or have abdominal gas. Moreover, it relies on an accurate robot-US calibration so the preview images correspond to the real ones. As mentioned in the following paragraphs, there are several techniques to mitigate this challenge. Most importantly, however, this method is intended primarily for fast initial navigation and rough alignment, after which the real image stream should be used for finding diagnostic images. Less accurate registration or more patient motion makes the preview less useful proportionally to the error, but does not eliminate the utility entirely. In the capacity of a rough initial positioning tool, as shown by the results, this method can reduce the expert's task load, increase the accuracy of sweeps, improve pressure application, and reduce completion time. These factors may contribute to making telerobotic ultrasound more practical and performant.

With this visual-haptic MMT concept, several other avenues of further research. For instance, in this system the expert is disconnected to a degree from the follower robot's actions, but the follower still carries out the expert's input trajectory. Instead, the expert and follower could be completely decoupled by making the robot follow the desired US image rather than the motion. More concretely, if the expert moves the haptic device and is presented with a certain image re-sliced from the sweep, the robot can be moved by visual servoing such that the real image matches the one seen by the expert as well as possible. In this way, the view of the expert is guaranteed to be accurate. However, this matching process is likely difficult to achieve in real time and could lead to unexpected robot motion. Additionally, this method does not guarantee that the expert is able to find a high-quality image unless the real US image is also shown, in which case the benefits may be marginal. However, similar approaches may constitute interesting future work.

Similarly, if the patient has moved, the anatomy in a new frame may be offset from the same anatomy in the existing sweep. Thus, the incoming image can be registered to the existing sweep to find the offset, and the sweep can be deformed accordingly. This could potentially compensate for small patient motions. In the case of a larger motion such as rolling onto one side for a kidney exam, however, the sweep would likely have to be recaptured. Furthermore, the US image depends also on the pressure applied by the transducer, which is not captured when simply re-slicing a sweep. Since the force is known on the haptic device, the local sweep can be deformed in real time to simulate the compression due to the pressure. This is already possible in the ImFusion Suite. Future work will also examine what level of motion, deformation, or registration error can be handled by such a system.

\bibliographystyle{ieeetr}
\bibliography{refs}
\end{document}